\documentclass[compsoc,journal]{IEEEtran}

\usepackage{amssymb,enumitem,subfigure}
\usepackage{soul}
\usepackage{url}
\usepackage{booktabs}       % professional-quality tables
\usepackage[table]{xcolor}
\usepackage{tcolorbox}
\usepackage[noadjust]{cite}
\usepackage[acronym]{glossaries}
\usepackage{flushend}

\newacronym[plural=VUEs,firstplural=vehicular users (VUEs)]{vue}{VUE}{vehicular user}

\newacronym[plural=RSUs,firstplural=roadside units (RSUs)]{rsu}{RSU}{roadside unit}
\newacronym{pdf}{PDF}{probability distribution function}

\newacronym{ar}{AR}{augmented reality}

\newacronym[longplural=base stations]{bs}{BS}{base station}

\newacronym{embb}{eMBB}{enhanced mobile broadband}

\newacronym{jnd}{JND}{just noticeable difference}

\newacronym{mmtc}{mMTC}{massive machine-type communication}

\newacronym{mmw}{mmWave}{milimeter wave}

\newacronym{mbs}{MBS}{macro base station}

\newacronym{lis}{LIS}{large intelligent surfaces}

\newacronym{los}{LoS}{line-of-sight}

\newacronym{ml}{ML}{machine learning}

\newacronym{nlos}{NLoS}{non-LoS}

\newacronym[\glslongpluralkey={radio frequencies}]{rf}{RF}{radio frequency}

\newacronym{rss}{RSS}{received signal strength}

\newacronym{sinr}{SINR}{signal-to-noise ratio}

\newacronym{urllc}{URLLC}{ultra-reliable and low-latency communication}

\newacronym{vr}{VR}{virtual reality}

\newacronym{aoi}{AoI}{age-of-information}

\newacronym{v2v}{V2V}{vehicle-to-vehicle}

\newacronym{tx}{TX}{transmitter}

\newacronym{rx}{RX}{receiver}

\newacronym{gpr}{GPR}{Gaussian process regression}

\newacronym{e2e}{E2E}{end-to-end}

\newacronym{qos}{QoS}{quality-of-service}

\newacronym{qoe}{QoE}{quality-of-experience}

\newacronym{nn}{NN}{neural network}

\newacronym{dnn}{DNN}{deep neural networks}

\newacronym{5g}{5G}{fifth generation}

\newacronym{mf}{MF}{mean-field}

\newacronym{ul}{UL}{upload}

\newacronym{dl}{DL}{download}

\newacronym{lidar}{LiDAR}{light detection and ranging}

\newacronym{fov}{FoV}{field-of-view}

\newacronym{tti}{TTI}{transmission time interval}

\newacronym{harq}{HARQ}{hybrid automatic repeat request}

\newacronym[\glslongpluralkey={radio access technologies}]{rat}{RAT}{radio access technolgy}

\newacronym{noma}{NOMA}{non-orthogonal multiple access}

\newacronym{tsn}{TSN}{time sensitive networking}

\newacronym{sota}{SoTA}{state-of-the-art}

\newacronym{mati}{MATI}{maximum allowable transfer interval}

\newacronym{mad}{MAD}{maximally allowable delay}

\newacronym{sati}{SATI}{stochastic maximum allowable transfer interval}

\newacronym{wncs}{WNCS}{wirelss networked control systems}

\newacronym{admm}{ADMM}{alternating direction method of multipliers}

\newacronym{uav}{UAV}{unmanned aerial vehicle}

\newacronym{hri}{HRI}{human-robot interaction}

\newacronym{rri}{RRI}{robot-robot interaction}

\newacronym{ccdf}{CCDF}{complementary cumulative distribution function}

\newacronym{rgbd}{RGB-D}{color and depth}

\newacronym{ris}{RIS}{reconfigurable intelligent surface}

\newacronym{cococo}{CoCoCo}{communication and control co-design}

\newacronym{prt}{PRT}{perception-reaction time}

\newacronym{rb}{RB}{resource block}

\newacronym{sl}{SL}{split learning}

\newacronym{ee}{EE}{energy efficiency}

\newacronym{snr}{SNR}{signal-to-noise ratio}

\newacronym{irt}{IRT}{isochronous real time}

\newacronym{fl}{FL}{federated learning}

\newacronym{pca}{PCA}{principal component analysis}

\newacronym{bibo}{BIBO}{bounded-input, bounded-output}

\definecolor{aliceblue}{rgb}{0.94, 0.97, 1.0}

\def\eg{e.g.,}

\title{\fontsize{21}{22}\selectfont When Wireless Communications Meet Computer Vision in Beyond 5G}% \textcolor{red}{[this is only a tentative title..}]}

\author{Takayuki~Nishio, Yusuke Koda$^\dagger$,
        ~Jihong~Park$^\ddag$,
        ~Mehdi~Bennis$^{\dagger\dagger}$,
        and~Klaus~Doppler$^{\ddag\ddag}$
\thanks{T.~Nishio is with the School of Engineering, Tokyo Institute of Technology, 152-8550, Tokyo, Japan. (email: nishio@ict.e.titech.ac.jp)}
\thanks{$^\dagger$Y. Koda is with the Graduate School of Informatics, Kyoto University, 606-8501 Kyoto, Japan (email: koda@imc.cce.i.kyoto-u.ac.jp).}
\thanks{$^\ddag$J.~Park is with the School of Information Technology, Deakin University, Geelong, VIC 3220, Australia (e-mail: jihong.park@deakin.edu.au).} 
\thanks{$^{\dagger\dagger}$ M.~Bennis is with the Centre for Wireless Communications, University of Oulu, 90014 Oulu, Finland (email: mehdi.bennis@oulu.fi). }
\thanks{$^{\ddag\ddag}$ K.~Doppler is with Nokia Bell Labs, Sunnyvale, CA 94086, USA. }
}

\begin{document}

\maketitle

\IEEEpeerreviewmaketitle

\glsresetall

\begin{abstract}
This article articulates the emerging paradigm, sitting at the confluence of computer vision and wireless communication, to enable beyond-5G/6G mission-critical applications (autonomous/remote-controlled vehicles, visuo-haptic VR, and other cyber-physical applications).
First, drawing on recent advances in machine learning and the availability of non-RF data, vision-aided wireless networks are shown to significantly enhance the reliability of wireless communication without sacrificing spectral efficiency. 
In particular, we demonstrate how computer vision enables {look-ahead} prediction in a millimeter-wave channel blockage scenario, before the blockage actually happens.
From a computer vision perspective, we highlight how radio frequency (RF) based sensing and imaging are instrumental in robustifying computer vision applications against occlusion and failure. 
This is corroborated via an RF-based image reconstruction use case, showcasing a receiver-side image failure correction resulting in reduced retransmission and latency.
Taken together, this article sheds light on the much-needed convergence of RF and non-RF modalities to enable ultra-reliable communication and truly intelligent 6G networks.

\end{abstract}

\glsresetall

%Section/subseciton number is not used in magazine.
\section*{Introduction}\label{sec:intro}
The overarching goal of ultra-reliable and low-latency communication (URLLC) lies in satisfying the stringent reliability and latency requirements of mission and safety-critical applications. In order to achieve these stringent requirements, current 5G URLLC solutions come at the cost of low spectral efficiency due to channel probing and estimation. In addition, 5G URLLC presumes a static channel model that fails to capture non-stationary channel dynamics and exogenous uncertainties (e.g., out-of-distribution or other under-modeled rare events), which are germane to uncontrolled environments~\cite{park2020}. %, swamy2019monitoring}. 
To overcome these fundamental limitations, driven by the recent advances in machine learning (ML) and computer vision, one key enabler for beyond-5G URLLC is leveraging visual data (\eg{} RGB depth (RGB-D) camera imagery, LiDAR point cloud, etc.) generated from a variety of vision sensors that are prevalent in intelligent machines such as robots, drones, and autonomous vehicles. From a wireless standpoint, these visual data enable a more accurate prediction of wireless channel dynamics such as future received power and channel blockages, as well as constructing high-definition 3D environmental maps for improved indoor positioning and navigation \cite{Nandakumar2018}. 
This line of works is referred to as \emph{view to communicate} (V2C).

% Furthermore, intelligent machines such as robots, drones, and autonomous vehicles are equipped with multiple vision sensors to perceive their surrounding environments by processing visual imagery and point clouds.
% From a wireless standpoint, visual data is also a key for understanding wireless channel dynamics, while enabling accurate prediction of future received power, channel blockages, and other important system parameters.

\begin{figure}[!t]
    \centering
    	\includegraphics[width=\columnwidth]{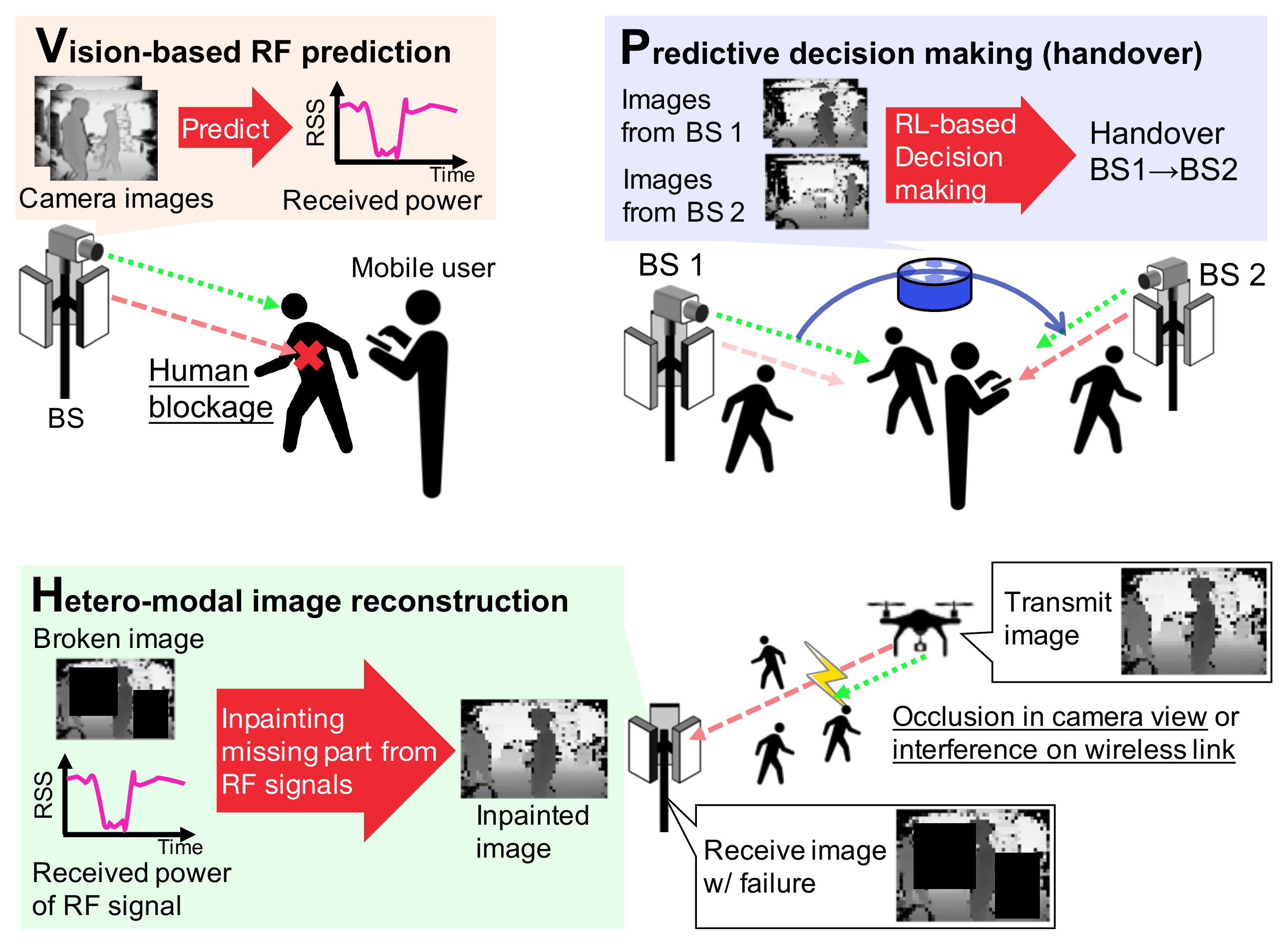} 
    \caption{An illustration of vision aided wireless communication, i.e., \emph{view to communicate (V2C)}, for millimeter-wave (mmWave) channel prediction and predictive handover, and RF signal assisted imaging, i.e., \emph{communicate to view (C2V)} for image inpainting.}
    \label{fig:teaser}
\end{figure}

On the other hand, in some scenarios computer vision is vulnerable to occlusions of visible light by walls, human body, and other environmental artifacts such as lighting. This can be addressed by leveraging radio frequency (RF) sensing such as using Wi-Fi signals to diffract and detour blockages as opposed to visible light, thereby precisely tracking user locations even behind walls \cite{Alahi:2015aa}. More recently, exploiting millimeter-wave (mmWave) and terahertz (THz) signals can provide even higher-resolution sensing capabilities that can penetrate body tissues for non-invasive medical imaging \cite{Doddalla2018}. This research direction is referred to as \emph{communicate to view} (C2V).

Motivated by the aforementioned confluence of computer vision and RF-based wireless transmission, this article sheds light on the synergies and complementarities of the integration of both visual and RF modalities for enabling URLLC in 5G and beyond. To this end, we discuss the challenges and research opportunities in V2C and C2V. Then, their feasibility is demonstrated using selected use cases, ranging from vision-aided mmWave channel prediction and proactive handover decisions, to image reconstruction of lost visual parts caused by packet loss. Finally, we conclude this article by laying down future research directions.

\section*{V2C: Vision-Aided Wireless Systems}
\label{sec:VAW}
A new paradigm in beyond-5G wireless systems is to leverage non-RF data, among which visual images complement traditional RF-based systems \cite{park2020}. For instance, one can predict future mmWave channel conditions using a sequence of camera images containing mobile blockage patterns \cite{Nishio2019}, thereby enabling proactive decision-making (\eg{} handover, beamforming, multi-path transmission, etc.). In what follows, the rationale related works, and future research opportunities of V2C are elaborated.

\subsection*{Vision-Based RF Channel Prediction}

%\tred{[JH: Challenges may be confused between (1) the challenges when applying our proposed method and (2) the challenges that motivate our proposed method. I think we focus mostly the latter. In order to debunk any misunderstanding, I changed Challenges into Motivation throughout the paper. For some sections, you may need to revise the descriptions accordingly.]}

\vspace{5pt}\noindent\textbf{Motivation.}\quad 
In beyond-5G systems, mmWave and THz signals are envisaged to play an important role thanks to their abundant bandwidth. However, these signals are highly directional and vulnerable to blockages, such as moving pedestrians, vehicles, and so forth. Hence, predicting the occurrence of blocked and non-blocked channels, that is line-of-sight (LOS) and non-LOS (NLOS), is crucial in ensuring reliable connectivity, notably for mission-critical applications. Predicting such events using past RF signals is extremely challenging while consuming spectral resources. To obviate this problem, visual data such as RGB-D images and 3D point cloud that capture a variety of hidden features in wireless environments (e.g., object locations, shapes, materials, and mobility patterns) can be exploited. In so doing, one can accurately predict future mmWave and THz channel conditions without consuming RF resources to probe and estimate the channels.

% expected to help address the bandwidth shortage problem. However, high frequency signals can be attenuated largely when the LOS path is blocked by obstacles such as pedestrians and vehicles. 

% In this regard, additional information regarding obstacles and reflectors can be obtained from visual data to improve the prediction accuracy of wireless signals . 

% The benefits of vision-based RF prediction are (1) accurate \textit{look-ahead} forecasting of moving obstacles and (2) obviating the need for RF resources for probing and channel estimation. %\tred{[JH: This last part is interesting, but is about the benefit of using RF data, not vision. This direction is the opposite to this section's scope.]} Moreover, RF prediction dataset can be constructed automatically while transmitting frames in the wireless system. Thus, we can generate large-volume dataset much easier than conventional vision problems, which improves prediction accuracy.

\vspace{5pt}\noindent\textbf{Related Works.}\quad
RGB-D images are useful for accurately predicting the future received power in mmWave (i.e., above 6 GHz) and sub-6 GHz carrier frequencies. In~\cite{Nishio2019}, the future mmWave received power is predicted by feeding past RGB-D images into a deep neural network (DNN), in which two randomly moving people block the communication link in an indoor experiment. 
Similarly, in~\cite{Ayva2019}, future 2.4 GHz channel states in an indoor experiment are accurately predicted using RGB-D images fed into a DNN. As demonstrated by these prior experiments, vision-based solutions can achieve accurate RF channel prediction without consuming any RF resources. This is in stark contrast to traditional channel prediction methods that frequently exchange RF pilot signals for high prediction reliability, which is not feasible in URLLC due to the stringent latency requirements.

\vspace{5pt}\noindent\textbf{Opportunities.}\quad
Beyond the aforementioned received power prediction, V2C has far more potential in predicting packet error rates, the number of reflectively propagating paths, optimal beam directions, to mention a few. Furthermore, in addition to indoor environments, it is worth studying the effectiveness of V2C in urban outdoor environments wherein the channel prediction becomes more challenging due to highly dynamic mobile blockage patterns, higher number of blockers and reflectively propagating paths. Last but not least, it is important to develop sample-efficient prediction techniques since conventional DNN training frameworks often require a large number of data samples. Alternatively, by exploiting meta learning and transfer learning, one can pre-train a DNN using easily accessible data (\eg{} data collected from public repository, ray-tracing simulations, etc.), and then fine-tune the DNN with only a few on-site data.

\subsection*{Hetero-Modal Vision-Based RF Channel Prediction}\label{sec:hetero-modal}
\vspace{5pt}\noindent\textbf{Motivation.}\quad
Fusing visual data with other modalities can enrich the useful features of wireless environments while complementing the missing features in the visual modality. Because vision is vulnerable to object occlusion while having restricted field-of-views (FoVs), audio data can partly complement such limitations; for example, by hearing the Doppler effect, one can predict a vehicle's moving direction and speed. 
%Klaus: I would replace audio with radar.
Another example is inertial measurement unit (IMU) data tracking the user movements during blockages which can also be used to track relative velocities to blockages.

% A promising approach to improve prediction accuracy is to leverage multi-modalities (\eg{} RF signals, RGB-D cameras, LiDARs, Gyroscope, and acceleration sensors). For example, acceleration information of mobile devices enables accurate prediction of devices' movements, resulting in improved RF channel prediction.

%Existing vision-based RF-prediction methods assume that the mobile devices are in the LOS of vision sensors and the vision is not largely blocked by obstacles while blockage occurs on the wireless propagation path. However, the mobile devices can exist in the NLOS of the vision sensor, and the occlusions could occur frequently especially in dense environments. 

\vspace{5pt}\noindent\textbf{Related Works.}\quad
DNNs are capable of fusing heterogeneous data modalities. In \cite{Ding2015}, 2D images and 3D face renderings are vectorized, and concatenated at the input layer of a convolutional neural network (CNN) for learning facial representations. 
In \cite{Koda:2019ac}, to predict future channel conditions, received RF signal power and RGB-D images are fused using a multi-modal split learning architecture, while RGB-D images captured from different FoVs are integrated via an average pooling layer. Such fusion can be immediately achieved without incurring any extra latency, as opposed to traditional data fusion algorithms consuming non-negligible computing time.

% DNN has been applied to make inference from multi-modalities. A deep learning framework to jointly learn face representation using RGB image and 3D face model was studied in . In \cite{Koda:2019ac}, multi-modal split learning was investigated, leveraging RF signals in addition to images for improving the accuracy of the vision-based received power prediction. The detail of this work is described in Sect.~\ref{sec:use_case} as a selected use case.

%split the neural network into multiple subnetworks and each subnetwork is handled with the device that has a sensor generating data for the subnetwork. The detail of this work is described in Sect.~\ref{sec:use_case} as a selected use case.

\vspace{5pt}\noindent\textbf{Opportunities.}\quad
Understanding the pros and cons of each data modality is crucial. As an example, for user localization, Wi-Fi signals (sub 6 GHz) are useful to cope with blockages \cite{Adib2013}, yet can hardly achieve high precision since the blockages result in NLOS communications.%due to the long wave lengths - Klaus: it is not the wavelength but the bandwidth that is important - Wi-Fi with 80MHz and 160MHz can achieve high accuracy (cm) because it can effectively separate multi-path. However, network based localization needs at least 2-3 AP in LOS.  
By contrast, mmWave signals (above 6 GHz) are vulnerable to blockages and require denser deployment. %Klaus: I added denser deployment because having LOS makes it more accurate.
Therefore, it is mostly LOS communications and can achieve high-precision localization. This highlights the importance of selecting and matching useful data types. Furthermore, the accuracy and cost of channel prediction hinge on how to fuse the hetero-modal data. For instance, compared to average pooling, concatenation consumes more energy due to the increase in the model size, in return for achieving higher prediction accuracy. Hence, it is important to optimize the fusion framework, consisting of DNN architectures, training algorithms, and data pre/post-processing, subject to energy requirements. To this end, split learning is a promising framework, in which a DNN is split into multiple subnetworks that are individually stored by each device \cite{Koda:2019ac}. 
By adjusting the cut layer, one can reliably satisfy each device's energy constraint.

\subsection*{Vision-Based Proactive Decision-Making}\label{sec:decision_making}
\vspace{5pt}\noindent\textbf{Motivation.}\quad
 Based on predicted future channel information, one promising way is to proactively carry out decision-making in wireless systems. For example, by predicting future blockage occurrences at each base station (BS), one can seamlessly handover users in order not to experience NLOS channels. In a similar vein, content can be proactively cached at the user prior to a blockage. %\textcolor{red}{However, since multi-node connectivity and content caching will consume more resources, one should use them carefully only when needed. Hence, accurate proactive decision-making is necessary.}

% . Predictive decision making emerges as a new challenge, whereby beamforming and handover are conducted predictively based on visual information for mitigating and avoiding future link blockage.

\vspace{5pt}\noindent\textbf{Related Works.}\quad
The effectiveness of predictive beamforming has been demonstrated in \cite{Iimori2020}. Therein, blockage patterns are learned by a statistical learning method, and a proactive beamforming algorithm is applied to reduce the link outage probability. Moreover, vision-based handover methods have been investigated in \cite{Koda:2019ab}, in which a reinforcement learning (RL) framework learns an optimal mapping from visual data onto handover strategies. More details of this work will be elaborated in a later section as a selected use case.
%in Sect.~\ref{sec:use_case} as a selected use case.
% [nishio] We cannot use section/subsection numbers in the magazine.

\vspace{5pt}\noindent\textbf{Opportunities.}\quad
Towards supporting URLLC, a single proactive decision based on a wrong prediction may cause catastrophic consequences. 
This calls for designing robustness against prediction failures and for increased prediction accuracy. For example, prediction errors due to video packet loss, dead camera pixels or stand vision sensors can be reduced by using RF data to reconstruct the distorted visual data, emphasizing the importance of the research direction of C2V, to be discussed in the following section.
%This calls for designing robust measures for prediction failures, for example, prioritizing the resource allocation.% \textcolor{red}{to be revised} Besides, for higher prediction accuracy, one can reconstruct distorted visual data (e.g., due to packet loss, dead camera pixels, stained vision sensors, etc.) using RF data, mandating the research direction of C2V, to be discussed in the following section.

% Performance degradation induced by prediction failure is a critical issue of predictive decision making. Despite substantial advancement in ML technology, prediction failures cannot be completely prevented, in which prediction failure could cause performance degradation. This calls for control mechanisms that are robust against prediction failures. 

\section*{C2V: Wireless-Aided Computer Vision}
\label{sec:RF-CV}

Traditional computer vision is based on the imagery captured using visible light, so is limited within line-of-sight (LOS). Compared to visible light, RF signals are more diffractive, thereby enabling non-LOS imaging (e.g., see-through-walls \cite{Adib2013, Doddalla2018}) that is necessary for non-intrusive inspection in mission-critical and time-sensitive applications. Furthermore, fusing the visible imagery and RF signals, one can improve the imagery resolution while reconstructing distorted or occluded objects. From the perspective of such a C2V research direction, the rationale, related works, and future opportunities are elaborated next.

% Radar is a well-known technology providing computer vision using RF signals.
% RF signals contain geometrical information in the environment but less information about identifying or exposing sensitive data, thereby RF-based vision is privacy-preserving, which is a challenge for conventional image-based computer vision.
% Moreover, recent developments in RF-based imaging are enabling to see behind objects and walls, that is NLOS imaging, which cannot be achieved with conventional camera-based imaging \cite{Adib2013, Doddalla2018}. 
% %NLOS imaging is an essential technology for mission-critical applications such as safe self-driving and smart rescue.
% NLOS imaging is an essential technology for mission-critical applications and helpful to prevent wrong predictions.
%The new challenges and opportunities motivated by these state-of-the-art technologies are discussed in the following sections. 

\subsection*{RF-Based Imaging}
\vspace{5pt}\noindent\textbf{Motivation.}\quad
Ultra-high frequencies such as mmWave and THz bands are expected to be a key enabler for high-resolution NLOS imaging.
Building walls and floors typically behave to a first order as mirrors and reflect the high-frequency signals, especially THz signals, which enables seeing behind walls %Klaus: through walls is misleading here
and around %Klaus: I added around because that is the main mechanism
corners assuming sufficient reflection or scattering paths \cite{rappaport2019}. 
In addition to NLOS imaging, mmWave and THz based imaging are less impacted by weather and ambient light compared to optical cameras.
Another advantage is short exposure time. The typical exposure time of optical camera is several to few tens milliseconds, while that of RF-imaging is microseconds, which enables high speed RF cameras to track fast movement.

\vspace{5pt}\noindent\textbf{Related Works.}\quad
The feasibility of THz-based NLOS imaging was demonstrated through imaging examples in the 220--330 GHz band using common building materials \cite{Doddalla2018}. A mmWave-based gait recognition method was studied for recognizing persons from their walking postures, which is expected to be still effective under non-line-of-sight scenarios \cite{meng2020}.

\vspace{5pt}\noindent\textbf{Opportunities.}\quad
Severe signal attenuation of mmWave and THz signals induced by pathloss and blockage limits the coverage of RF-based imaging. 
In 5G/6G networks based on mmWave and THz bands, highly directional antenna and densely deployment are exploited to compensate the signal attenuation, and hence these solutions could be utilized in mmWave/THz-based imaging. However, interference among both RF-based imaging and mmWave/THz communication systems remains a critical issue. 
For enabling co-existence of RF-based imaging and communication systems, one can exploit wireless resource scheduling and multiple access mechanisms such as time division multiple access (TDMA), carrier sense multiple access/collision avoidance (CSMA/CA), and non-orthogonal multiple access (NOMA). Another way to mitigate the interference is interference cancellation which processes the known transmitted imaging or communication signal to generate a negative that, when added to the composite signal, reverts the effect of the interference.

\subsection*{Multi-Band RF-Based Imaging}
\vspace{5pt}\noindent\textbf{Motivation.}\quad
Improving resolution and accuracy is an important challenge in RF-based imaging. To this end, joint use of multiple signals on different frequency band is a promising way. Recent wireless networks can leverage multiple frequency bands. For example, Wi-Fi devices will be able to utilize sub-GHz, 6 GHz, and mmWave (60 GHz) in addition to 2.4 and 5 GHz. Such different frequencies have different propagation characteristics, resulting in different resolution and FoV on imaging. Thus, cooperatively using multiple signals could improve image resolution and sensing accuracy.

\vspace{5pt}\noindent\textbf{Related Works.}\quad
A super-resolution of multi-band radar data on 3--12\,GHz bands and decimeter-level localization leveraging multi-band signals on 900\,MHz, 2.4\,GHz, and 5\,GHz have been studied in \cite{Zhang2014} and \cite{Nandakumar2018}, respectively. These works demonstrated that cooperative use of multiple signals on different frequency bands can improve imaging resolution or localization accuracy.

\vspace{5pt}\noindent\textbf{Opportunities.}\quad
With increase in range of frequency bands (\eg{} joint use of sub-6\,GHz and mmWave signals), we need to consider resolution-coverage trade-off.
MmWave and THz signals enable high-resolution imaging, but the high attenuation limits the coverage of RF-based imaging. On the other hand, lower frequency (sub-GHz) generates lower-resolution images than mmWave/THz imaging, but their coverage is wider than mmWave/THz imaging. 
Therefore, adaptive use of multiple frequency bands is expected to achieve better trade-off between the resolution and coverage.
Moreover, utilizing multiple channels and wider bandwidth could cause severe interference with multiple communication systems and make the interference management more difficult. Thus, the co-existence mechanism of communication and imaging systems becomes more important.

\subsection*{Hetero-modal RF-Based Imaging}\label{sec:multi_modal}
\vspace{5pt}\noindent\textbf{Motivation.}\quad
Leveraging heterogeneous modalities could be another solution to improve resolution and reliability of imaging. Smart devices such as smart phones, vehicles, and drones have multiple imaging sensors (\eg{} camera and LiDAR) and RF modules (\eg{} Wi-Fi, Bluetooth, 4G/5G, and WiGig). We can exploit these modalities cooperatively for imaging and sensing. However, there is an open issue of how to integrate the heterogeneous modalities.

\vspace{5pt}\noindent\textbf{Related Works.}\quad
In computer vision, deep learning based multi-modal image fusion is studied for improving image quality \cite{LIU2018}. 
Multi-modal images (\eg{} Visual, IR, CT, and MRI images) are fused based on their pixel values via some fusion rule, which is called as pixel level fusion.
There are other fusion approaches; feature level fusion and decision level fusion. In the feature level fusion, prominent features (\eg{} edges, corner points, and shapes) are extracted from different images and combined into a feature map. In the decision level fusion, the different images are pre-processed and leveraged for decision making separately. Then, the individual decisions are integrated to provide more accurate decision.

\vspace{5pt}\noindent\textbf{Opportunities.}\quad
Although the pixel level fusion generally requires a heavier computation than other level fusion techniques, it is still widely used in many fields such as remote sensing because of higher accuracy. A major issue of the multi-modal imaging is spatial and temporal misregistration induced by different image scales, resolutions, and deployed angles and locations of the sensors.
Moreover, in RF-based imaging, there could be a new fusion level, that is the signal level fusion. In the fusion, RF signals are directly fused, and new features are generated for more accurate imaging. %Sect.~\ref{sec:use_case3} 
The next section details a case of the signal level fusion for predicting a missing part of an image.

\section*{Selected Use Cases}\label{sec:use_case}
\subsection*{Hetero-Modal mmWave Received Power Prediction}\label{sec:use_case1}
As discussed in the earlier section, %Sect.~\ref{sec:hetero-modal}, 
past image sequences are informative to forecast sudden LOS and NLOS transitions, which is hardly observable from RF received power sequences.
On the contrary, past RF received power sequences are informative to predict future received powers highly correlated with the past ones under LOS conditions.
To benefit from these two modalities and thereby achieve better accuracy, the prediction method fusing these two modalities are studied as follows.

\vspace{3pt}\noindent\textbf{Scenario.}\quad 
Consider a depth camera with 30\,Hz frame rate monitoring a mmWave link that is intermittently blocked by two moving pedestrians.
Our objective is to predict future received powers with a look ahead horizon of 120\,ms based on a past depth image sequence and a received power sequence.
To this end, a split NN architecture is designed to integrate the depth image and received power sequences, thereby performing mmWave received power prediction with the two types of modalities.
Specifically, the split NN comprises convolutional layers that extract image features and a recurrent layer that concatenates the sequence of image features and RF received powers and performs time-series prediction of mmWave received power\,\cite{Koda:2019ab}.

\begin{figure}[t]
    \centering
    	\includegraphics[width=\columnwidth]{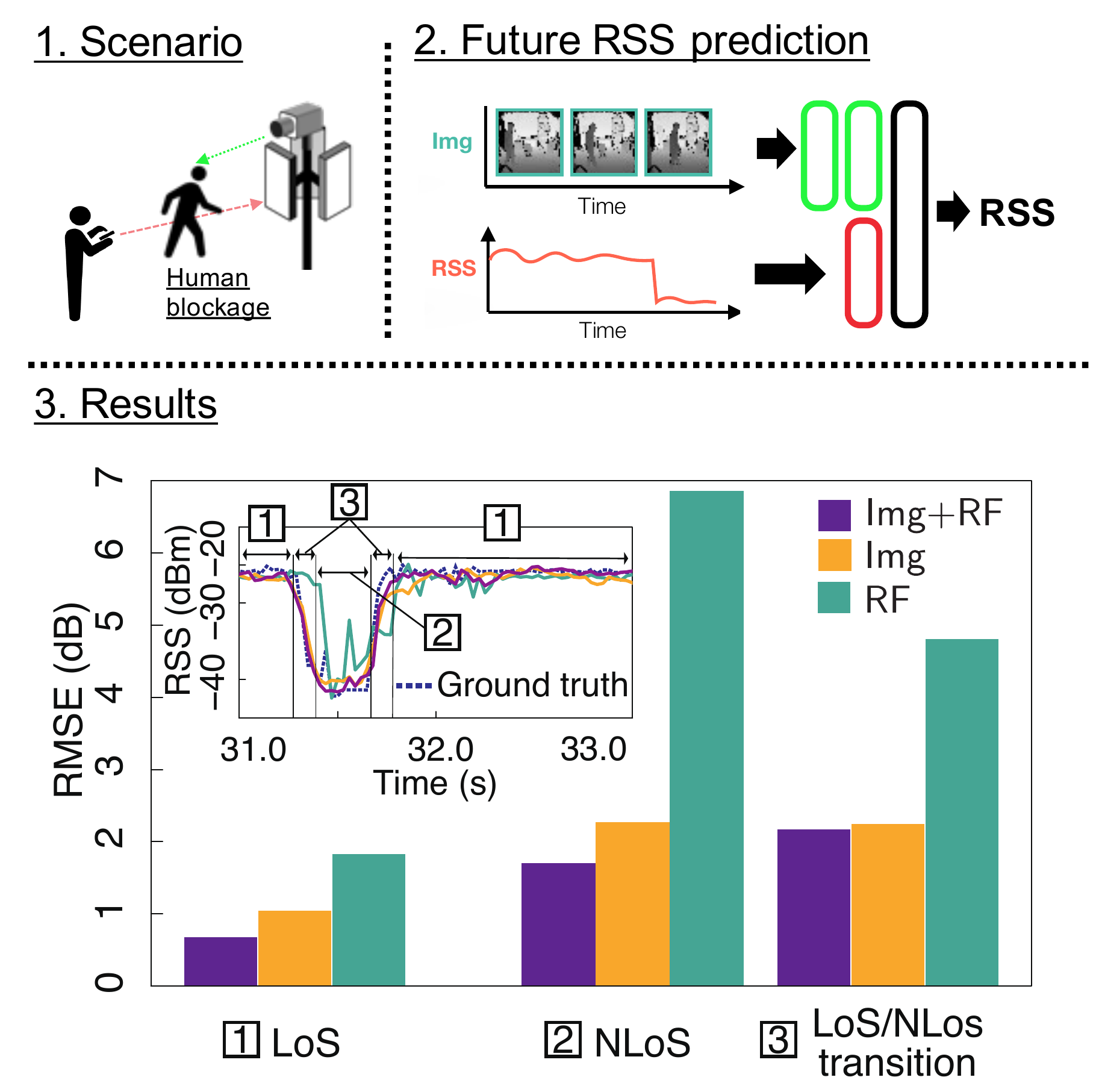} 
    \caption{Hetero-modal mmWave received power prediction. The error between actual \gls{mmw} \gls{rss} and its predicted value by fusing past \gls{rss} data and \gls{rgbd} images. %in different channel conditions.
    }
    \label{fig:use_case_non_rf}
\end{figure}

\vspace{3pt}\noindent\textbf{Results.}\quad
In Fig.~\ref{fig:use_case_non_rf}, showing the prediction accuracy in root-mean-square error (RMSE) in different channel conditions, we demonstrate that the prediction using both images and RF received powers (\textsf{Img+RF}) achieves higher prediction accuracy than the prediction using either one (\textsf{Img} and \textsf{RF}).
\textsf{Img+RF} does not only predict LOS/NLOS transitions as well as \textsf{Img}, but also predicts received powers correlated with the input received power sequence for a given LOS and NLOS conditions better than \textsf{Img}. 
This result exactly demonstrates the feasibility of the benefit from integrating image and RF modalities.

\subsection*{Multi-Vision Based Predictive mmWave BS Handover}
This section introduces the use case of handover management in mmWave communications to illustrate the importance of vision-based proactive decision-making.

\vspace{3pt} \noindent \textbf{Scenario.} \quad
%Handover always incurs a latency,
%Klaus: 5G Rel. 16 has make before break handover (dual active protocol stack) and does not incur latency. Therefore, I would rephrase this with Handovers may result in disruptions of the communication link
Handover can result in disruptions of the communication link, and hence, in determining handover timings, one should be aware of not only current RF conditions, but also how well each BS performs in a long run to prevent myopic decisions.
Traditionally, handover strategy is formed based on current RF conditions (e.g., channel state or received power); however, RF conditions are not necessarily informative to forecast sudden transitions between LOS and NLOS conditions.
This is where image modality comes as a rescue, wherein one can form a handover strategy being aware of mobility of obstacles and thereby predicting future LOS and NLOS transitions \cite{Koda:2019ac}.
Moreover, recent advancement of RL, namely deep RL, helps us achieve the aforementioned objective by feasibly handling a higher dimensionality of image modalities.

For example, in a simple scenario with two static BSs and single static STA.
The BS serving higher received power in LoS conditions (termed BS~1) initially associates with the STA.
The other BS termed BS~2 is a candidate to which a handover is performed and thereby compensates for the degraded data rates provided by BS~1 due to LOS blockages.
The BSs are served by depth cameras with depth images. 
The associated BS runs deep RL to learn the optimal \textit{action value} associated with each depth image that quantifies how well each BS performs in a long time horizon.
Thereby, the optimal timing to perform a handover is determined.

\begin{figure}
    \centering
    \includegraphics[width=\columnwidth]{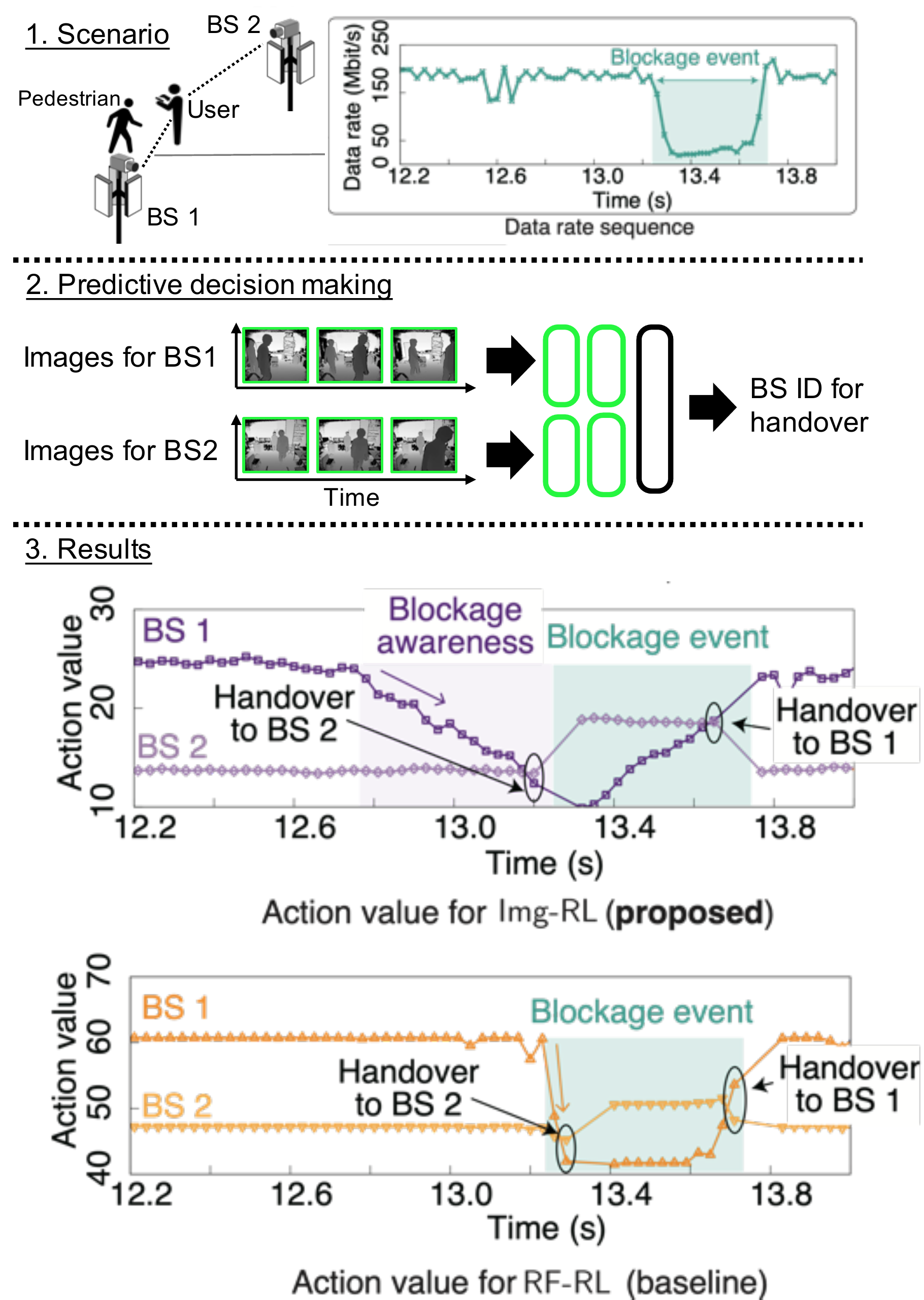}
    \caption{RL-based predictive handover. Action value for selecting each BS learned in \textsf{Img-RL} using depth images and in \textsf{RF-RL} using RF received powers.}
    \label{fig:DRL_HO}
\end{figure}
% The STA is initially associated with a BS observing higher receiver power from the STA.

\vspace{3pt} \noindent \textbf{Results.}\quad
Fig.~\ref{fig:DRL_HO} shows the learned action value using images (\textsf{Img-RL}) and RF received powers (\textsf{RF-RL}).
The action value for selecting BS~1 learned in \textsf{Img-RL} decreases as a pedestrian approaches a LoS path while that in \textsf{RF-RL} does not.
This result exactly indicates that \textsf{Img-RL} feasibly forms a handover policy being aware of future blockage events.
Therein, \textsf{Img-RL} triggers a handover earlier than \textsf{RF-RL}, and thereby, avoids the blockage event.
Thus, \textsf{Img-RL} exhibits a higher throughput (118\,Mbit/s) than \textsf{RF-RL} (113\,Mbit/s).

\subsection*{Hetero-Modal Image Reconstruction}\label{sec:use_case3}
This section presents the setting of mmWave signal-aided image reconstruction as a use case of hetero-modal RF-based imaging. %The goal is to solve an inverse problem of vision-assisted mmWave received power prediction, 
The goal of this work is image inpainting with hetero-modal information, in which a missing part of an image is reconstructed from the defective image and a sequence of mmWave received power values. 

\vspace{3pt} \noindent \textbf{Scenario.} \quad
Consider a depth camera monitoring a mmWave link that is intermittently blocked by two moving pedestrians, but a part of the image is missing due to occlusion or failure on the camera. 
The objective is to reconstruct the missing part of the image based on signal attenuation on the mmWave link. As shown in the previous use cases, the mmWave signals are strongly attenuated when an obstacle blocks LOS path, and the timing and intensity of the attenuation suggest where the obstacle moved.

Deep auto-encoder is leveraged for the hetero-modal inpainting. The encoder has two input layers for an image with occlusion and sequence of received power of RF signals. The input layer for image is followed by convolution layers, and the input layer for RF signal is followed by fully connected layers. The outputs of these layers are concatenated at the end of the encoder part and inputted to the decoder part consisting of convolution layers, which depicts an image including missing parts.

\begin{figure}
    \centering
    \includegraphics[width=\columnwidth]{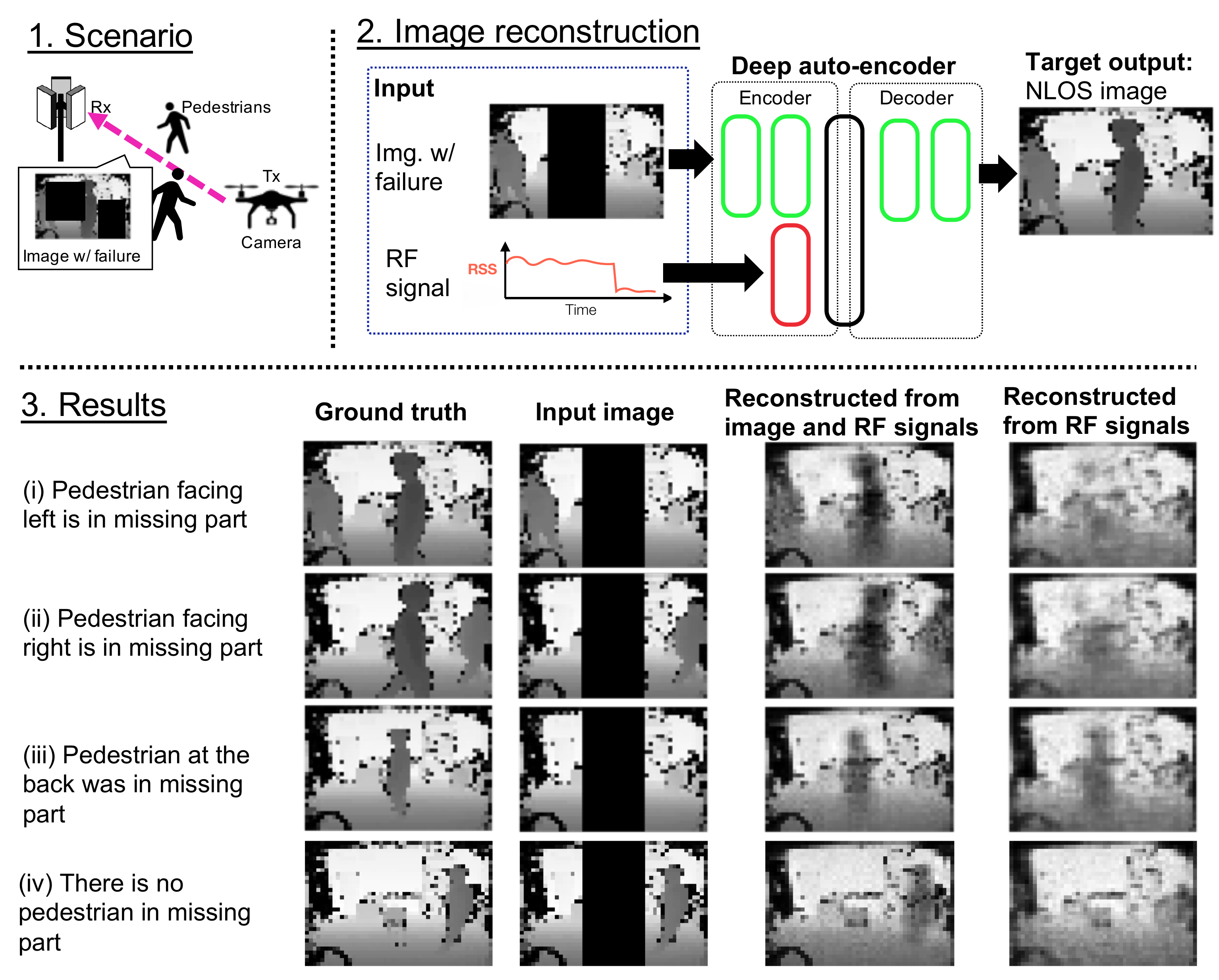}
    \caption{Hetero-modal image reconstruction. A missing part in the image is reconstructed from the received power of RF signals and an image with failure by deep auto-encoder.}
    \label{fig:NLOS_image}
\end{figure}

\vspace{3pt} \noindent \textbf{Results.}\quad
Fig.~\ref{fig:NLOS_image} depicts samples of depth-camera image without missing parts (ground truth), defective image (input image), and images reconstructed from both the input image and RF signal (\textsf{Img+RF}), or only from RF signal (\textsf{RF only}). 
Even though the reconstructed imaging uses limited features of RF signal, that is 32 points of mmWave received power sampled at 66\,ms intervals, \textsf{Img+RF} depicts the missing part on the input images as similar to the ground-truth images for both cases where the pedestrian was in the missing part or not.
Such inpainting of imagery with a large missing part is difficult for the conventional image inpainting which leverages only the imagery, because there is no information about the missing part. In contrast, the \textsf{Img+RF} reconstruction can obtain the information about the missing part from RF signals and reconstruct it.

Moreover, the \textsf{Img+RF} reconstruction depicts images more accurately than \textsf{RF only}. Thereby we can find even the direction of the pedestrian on the image reconstructed by \textsf{Img+RF}.
These results demonstrate the feasibility of image inpainting with RF signals and the benefit from integrating image and RF modalities.

\section*{Conclusions}\label{sec:conclusions} 
This article outlined the vision of fusing computer vision and wireless communication to spearhead the next generation of URLLC toward beyond-5G/6G mission-critical applications. This convergence opens up untapped research directions that go beyond the scope of this paper. An interesting direction is to investigate how much visual information is contained in RF signals of wireless communications. As demonstrated in the selected use cases, %Sect.~\ref{sec:multi_modal}, 
mmWave communication signals contain visual information of obstacles and help image inpainting. However, the current inpainted images are mimicry of training data, and it is still unclear what information (\eg{} shape and location of obstacles) can and cannot be retrieved from RF signals, calling for a novel visual information capacity analysis for a given task.
%\textcolor{red}{can add more futuristic threads here..}
%\textcolor{blue}{[how about this? nishio]
%Another interesting direction is to predict quality of experience (QoE) of mobile users for their wireless communications from computer vision. RGB cameras can capture faces and behaviors of mobile users, which may contain information about how they feel about their communication services.}
%\textcolor{blue}{[how about this? klaus]}
Another interesting direction is the creation and update of a real-time digital replica of the physical space around wireless access points using hetero-modal sensing that combines RGB-D, LiDAR, and RADAR. Such a replica can keep track of and predict the movement of people and objects through space. The resulting 3D model can also be utilized to perform ray tracing simulations to predict RF link quality. Moreover, RGB-D cameras can capture faces and behaviors of mobile users, through which their quality of experiences (QoEs) can be accurately predicted.

%\tred{[joint C2V-V2C resource allocation]}

% which may contain information about how they feel about their communication services. 

% the quality of experience (QoE) of mobile users can be accurately predicted 

\ifCLASSOPTIONcaptionsoff
  \newpage
\fi

\bibliographystyle{IEEEtran}
\bibliography{IEEEabrv,VAW} % should be less or equal to 15.

\begin{IEEEbiographynophoto}{Takayuki Nishio}
is an associate professor at the School of Engineering, Tokyo Institute of Technology, Japan.
He received the B.E.\ degree in electrical and electronic engineering and the master’s and Ph.D.\ degrees in informatics from Kyoto University in 2010, 2012, and 2013, respectively. 
He was an assistant professor in communications and computer engineering with the Graduate School of Informatics, Kyoto University from 2013 to 2020. From 2016 to 2017, he was a visiting researcher in Wireless Information Network Laboratory (WINLAB), Rutgers University, United States. His current research interests include machine learning-based network control, machine learning in wireless networks, and heterogeneous resource management.
\end{IEEEbiographynophoto}
\begin{IEEEbiographynophoto}{Yusuke Koda}
received the B.E. degree in electrical and electronic engineering from Kyoto University in 2016, and the M.E.\ degree from the Graduate School of Informatics, Kyoto University in 2018, where he is currently pursuing the Ph.D. degree. In 2019, he visited the Centre for Wireless Communications, University of Oulu, Finland, to conduct collaborative research. He received the VTS Japan Young Researcher’s Encouragement Award in 2017. He was a Recipient of the Nokia Foundation Centennial Scholarship in 2019.
\end{IEEEbiographynophoto}
\begin{IEEEbiographynophoto}{Jihong Park}
is a Lecturer at the School of IT, Deakin University, Australia. He received the B.S. and Ph.D. degrees from Yonsei University, Korea. His research interests include ultra-dense/ultra-reliable/mmWave system designs, as well as distributed learning/control/ledger technologies and their applications for beyond-5G/6G communication systems. He served as a Conference/Workshop Program Committee Member for IEEE GLOBECOM, ICC, and WCNC, as well as for NeurIPS, ICML, and IJCAI. He is an Associate Editor of Frontiers in Data Science for Communications, and a Review Editor of Frontiers in Aerial and Space Networks.
\end{IEEEbiographynophoto}
\begin{IEEEbiographynophoto}{Mehdi Bennis}
is an associate professor at the Centre for Wireless Communications, University of Oulu, Finland. His main research interests are in radio resource management, and machine learning in 5G/6G networks.
 He has coauthored 1 book and published more than 200 research articles in international conferences, journals, and book chapters. He was a recipient of the prestigious 2015 Fred W. Ellersick Prize from the IEEE Communications Society, the 2016 Best Tutorial Prize from the IEEE Communications Society, the 2017 EURASIP Best Paper Award for the Journal of Wireless Communications and Networks, and the all-University of Oulu Research Award.
\end{IEEEbiographynophoto}
\begin{IEEEbiographynophoto}{Klaus Doppler}
is the technical lead of the Mirror X project in Nokia Bell Labs. He has been heading the Indoor Networks Research focusing on ubiquitous Gigabit connectivity and platforms for smart buildings, enterprises and factories. In the past, he has been responsible for wireless research and standardization in Nokia Technologies, incubated a new business line and pioneered research on D2D Communications underlaying LTE networks. He received inventor awards in Nokia for 100+ granted patent applications. He has published 40+ scientific publications, received his PhD. from Aalto University, Finland in 2010 and his MSc. from Graz University of Technology, Austria in 2003.
\end{IEEEbiographynophoto}

%------------------------------------
\end{document}